\documentclass[12pt]{article}
\usepackage{amsfonts}
\usepackage{graphicx}
\usepackage{hyperref}
\usepackage{amsmath}
\usepackage{amssymb}
\usepackage{graphicx}
\usepackage{caption}
\usepackage{subcaption}
\usepackage{multirow}
\usepackage{color}

\begin{document}
\title{TRec: Learning Hand-Object Interactions through 2D Point Track Motion}
\author{Dennis Holzmann \thanks{Corresponding Author} \\ Bielefeld University, Faculty of technology, Germany \\ dennis.holzmann@uni-bielefeld.de \and Sven Wachsmuth \\ Bielefeld University, Faculty of technology, Germany \\ swachsmu@techfak.uni-bielefeld.de}
\date{}
\maketitle              
\begin{abstract}

We present a novel approach for hand-object action recognition that leverages 2D point tracks as an additional motion cue. 
While most existing methods rely on RGB appearance, human pose estimation, or their combination, 
our work demonstrates that tracking randomly sampled image points across video frames can substantially improve recognition accuracy. 
Unlike prior approaches, we do not detect hands, objects, or interaction regions. 
Instead, we employ CoTracker to follow a set of randomly initialized points through each video and use the resulting trajectories, 
together with the corresponding image frames, as input to a Transformer-based recognition model. 
Surprisingly, our method achieves notable gains even when only the initial frame and the point tracks are provided, 
without incorporating the full video sequence. 
Experimental results confirm that integrating 2D point tracks consistently enhances performance compared to the same model trained without motion information, 
highlighting their potential as a lightweight yet effective representation for hand-object action understanding.

\end{abstract}

\section{Introduction}

The task of action recognition involves identifying and classifying human actions from video data.
While many prior works focus on generic third-person datasets that capture large-scale body movements, 
our study targets hand-object interaction scenarios, where meaningful actions are characterized by fine-grained manipulations rather than full-body motion.
Such datasets can include both egocentric and third-person viewpoints, providing diverse perspectives on close-range interactions between hands and objects.
This perspective introduces unique challenges, including strong camera motion, frequent occlusions, and limited field of view, all of which complicate the modeling of both actor and object dynamics.

The prominent camera motion paired with the exceptional close and fast-moving background covering an extensive amount of the image pose an intricate challenge of modeling both actor and object dynamics.
The interplay of camera motion combined with the rapidly moving and closely captured background dominating the visual field make it particularly challenging to model both actor and object dynamics. 
Many existing approaches rely on RGB-based features. 
While such methods have achieved strong results, they often depend on computationally expensive or error-prone semantic detection pipelines for hands, objects, or skeletons. 

In this work, we investigate whether simple, semantically agnostic motion cues, specifically random 2D point tracks, can provide comparable or even superior information for recognizing hand-object interactions.

By tracking multiple points on hands, objects, and background regions, our method captures diverse motion signals, including camera movement and object deformation. 
The difficulty of distinguishing object categories is reduced, since our method directly tracks the motion of the object in the scene.
We generate these tracks using CoTracker.
The classification pipeline uses the ResNet18 as backbone and a Transformer based module on top for classification. 
Surprisingly, our experiments show that a Transformer model trained with such point tracks, even when only the initial frame is used, can recognize actions more accurately than the RGB-only baseline using the whole video.

Our main contributions are as follows:
\begin{itemize}
    \item We propose a simple and efficient method for hand-object action recognition that integrates 2D point tracks with image features.
    \item We demonstrate that these tracks capture meaningful motion patterns spanning hand, object, and head movements, without requiring explicit semantic detection.
    \item We show consistent performance improvements over the RGB-only baseline on an hand-object interaction benchmark.
\end{itemize}

\section{Related Work}

\textit{Trajectory-based methods} model motion by tracking key points or regions across consecutive frames \cite{wangActionRecognitionImproved2013,nguyenDenserTrajectoriesAnchor2018,yiMotionKeypointTrajectory2018}. 
They have been explored in exocentric (third-person) action recognition, where stable camera viewpoints simplify trajectory estimation. 
To the best of our knowledge, no prior work has investigated the use of 2D point tracks for hand-object action recognition without filtering background points, 
where strong camera motion and frequent occlusions make trajectory-based modeling more challenging.
Additionally, in contrast to these methods we utilize a relatively sparse set of points to track within the image compared to methods using optical flow approaches.
Wang \textit{et al.}~\cite{wangActionRecognitionImproved2013} introduced \textit{Improved Dense Trajectories}, 
which compensate for camera motion using RANSAC-based estimation and employ a human detector to distinguish foreground motion from background motion. 
This approach established a strong pre-deep-learning baseline for video action recognition. 
Nguyen \textit{et al.} \cite{nguyenDenserTrajectoriesAnchor2018} proposed \textit{Denser Trajectories of Anchor Points for Action Recognition}, 
which anchor trajectories to more stable points to better capture fine-grained motion variations and increase robustness to camera movement.

\textit{Hand-object detectors} can improve hand-object interaction recognition. 
Several methods utilize some kind of hand-object detector and integrate this information into their model \cite{chatterjeeOpeningVocabularyEgocentric2023a,shiotaEgocentricActionRecognition2024,peironeEgocentricZoneawareAction2024,materzynskaSomethingElseCompositionalAction2020}.
Chatterjee \textit{et al.}~\cite{chatterjeeOpeningVocabularyEgocentric2023a} expanded the vocabulary of egocentric actions by linking detected hands and objects with fine-grained action categories. 
Shiota \textit{et al.}~\cite{shiotaEgocentricActionRecognition2024} explicitly modeled hand-object contact and object state transitions, 
while Peirone \textit{et al.}~\cite{peironeEgocentricZoneawareAction2024} proposed a zone-aware recognition frameworks which learns environmental affordances to extract environment and action information separately to train an action recognition model. 
Materzyńska \textit{et al.}~\cite{materzynskaSomethingElseCompositionalAction2020} proposed \textit{Something-Else}, 
a compositional model that builds spatial-temporal interaction graphs between hands and objects, 
achieving strong performance on the Something-Something-v2 dataset.
Although effective, these methods depend heavily on accurate hand and object detection, which can fail under occlusions or strong motion.

\textit{RGB-based approaches}~\cite{sudhakaranLSTALongShortTerm2019,arnabViViTVideoVision2021a,liuVideoSwinTransformer2021,girdharAnticipativeVideoTransformer2021,tongVideoMAEMaskedAutoencoders2022a,wangInteractivePrototypeLearning2021} rely solely on raw visual appearance for action recognition, without incorporating explicit motion or geometric cues. 
Many of these high-performing models are trained on large-scale datasets, often leveraging extensive pretraining and carefully designed architectures. 
In contrast, we deliberately adopt a simple model design to isolate and clearly demonstrate the effectiveness of incorporating 2D point tracks. 
Unlike RGB-based features that primarily encode texture and color statistics, point trajectories provide explicit geometric and temporal information, capturing object and camera motion directly from the image sequence.
Girdhar \textit{et al.}~\cite{girdharAnticipativeVideoTransformer2021} propose the \textit{Anticipative Video Transformer}, which models temporal dependencies in egocentric videos to predict upcoming actions. 
Their model implicitly attends to image regions containing hands and objects, and its performance improves with increased temporal context. 
However, the paper does not clarify whether motion is inferred implicitly through appearance changes between frames. 
In contrast, our approach introduces motion information explicitly by integrating 2D point tracks, enabling the model to reason directly about the displacement and dynamics of scene elements rather than relying on learned appearance differences. 
This explicit representation of motion complements the visual features and enhances interpretability in modeling fine-grained hand-object interactions and providing robustness to camera motion.
Tong \textit{et al.}~\cite{tongVideoMAEMaskedAutoencoders2022a} employ masked autoencoder pretraining to improve video representation learning, achieving strong performance on egocentric and other video benchmarks. 
Wang \textit{et al.}~\cite{wangInteractivePrototypeLearning2021} further extend pretrained video backbone models through an interactive prototype learning framework, enhancing active-object representations by leveraging actor motion cues within RGB inputs.

\textit{Point tracking} aims to estimate the trajectories of 2D points across video frames, providing fine-grained motion information that supports higher-level tasks such as action recognition and object interaction analysis. 
Recent advances in this area have focused on improving both tracking accuracy and computational efficiency.
CoTracker~\cite{karaevCoTrackerItBetter2023} introduced a transformer-based architecture capable of jointly tracking multiple points across frames. 
Subsequent versions have further improved robustness, accuracy, and inference speed, establishing CoTracker as a state-of-the-art solution for general-purpose point tracking. 
In our work, we adopt CoTracker due to the public availability of pretrained weights, which allows for efficient and reliable integration of high-quality point trajectories into our framework.
EgoPoints~\cite{egopoints2024} proposed a dedicated benchmark for dense point tracking in egocentric videos. 
They specifically addresse the challenges of egocentric settings, such as strong camera motion and frequent occlusions, and provide a comprehensive evaluation protocol for tracking performance. 
In addition, the authors improve upon CoTracker by adapting it to egocentric motion patterns and by providing a large-scale dataset for benchmarking.

\section{Action Recognition using Point Tracks}

We present \textbf{TRec}, a transformer-based model for hand-object interaction recognition that jointly leverages 2D point tracks and image representations. 
TRec tracks both background and foreground points throughout a video to capture rich motion cues without relying on explicit object or hand detections. 
The model employs a lightweight image encoder to extract spatial features from individual frames and a transformer architecture~\cite{vaswaniAttentionAllYou2023a} to integrate temporal dependencies across frames. 
A set of randomly sampled 2D points is initialized within each video, covering both foreground and background regions, and their trajectories are tracked over time. 
The transformer receives as input the encoded image features together with these 2D point tracks, enabling it to reason jointly about appearance and motion information. 
An additional multi-headed attention layer is used to aggregate the transformer outputs, and the resulting representation is passed through an multi-layer perceptron (MLP) based action recognition head to predict the action depicted in the video. 
We evaluate TRec on the Something-Something-v2~\cite{goyalSomethingSomethingVideo2017} dataset.

In the following, we first describe the dataset curation process, including sampling strategies and preprocessing steps used to generate point trajectories. 
Next, we present the point tracking procedure and explain how trajectories are integrated with image features. 
We then detail the overall model architecture, including the image encoder, transformer design, and the action recognition head. 
Finally, we outline the training setup and evaluation protocol used to assess TRec on egocentric benchmarks.

\subsection{Dataset Curation and Preprocessing}

For training and evaluation, we use the Something-Something-v2 dataset~\cite{goyalSomethingSomethingVideo2017}.
We employ CoTracker~\cite{karaevCoTrackerItBetter2023} to obtain 2D point trajectories across video frames.

\textit{Something-Something-v2.}  
The Something-Something-v2 dataset~\cite{goyalSomethingSomethingVideo2017} contains short video clips depicting human-object interactions recorded in a wide variety of everyday environments.  
It comprises 174 action templates describing generic activities of someone doing "something”, such as pushing, pulling, or placing objects.  
It offers greater diversity in both visual context and action types, as well as a substantially large number of annotated samples, making it a valuable benchmark for training data-driven models.  
The dataset contains a diverse collection of videos, including egocentric recordings with significant camera motion as well as scenes captured from mostly stationary viewpoints. 
We use the dataset in its original form without any additional preprocessing or modifications, applying only standard data augmentation techniques.

\begin{figure}[ht]
    \centering
    \begin{subfigure}{0.495\textwidth}
        \includegraphics[width=\hsize]{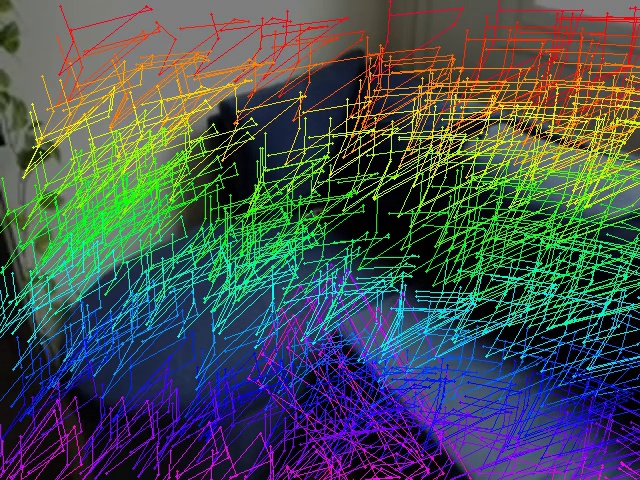}
        \caption{Throwing CD cover against sofa }
    \end{subfigure}
\hfill
    \begin{subfigure}{0.495\textwidth}
        \includegraphics[width=\hsize]{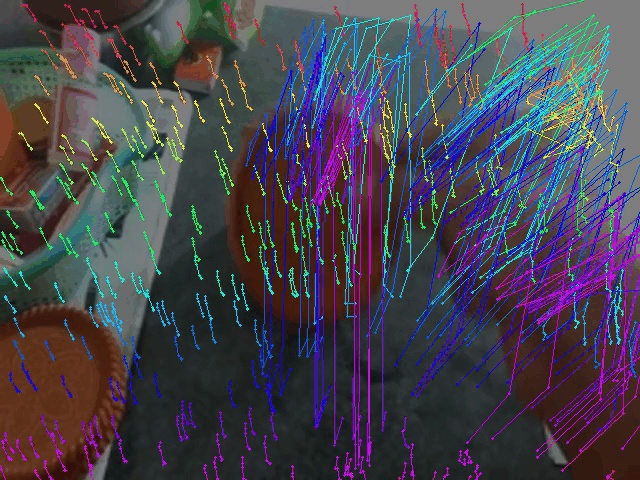}
        \caption{Put coaster into coaster holder }
    \end{subfigure}
    \caption{Trajectory of 2D points in example videos of the Something-Something dataset}
    \label{fig:point_sample}
\end{figure}

\textit{Point Track Generation.} 
We curate 2D point trajectories using CoTracker3~\cite{karaevCoTracker3SimplerBetter2024}. 
Figure~\ref{fig:point_sample} illustrates several examples of the generated point tracks.
For each action sequence, we randomly sample 900 points covering the entire image, including both foreground and background regions. 
Video frames are sampled at 30~fps to obtain smooth trajectories. 
The resulting point coordinates are normalized following the procedure described in~\cite{bharadhwajTrack2ActPredictingPoint2024}.
For each video we track a set of 900 points $p^i = \{(x^i_t, y^i_t) \in \mathbb{R}^{2} | t=0,...T\} $.
Here, $i \in [0, P]$ denotes the $i^{\text{th}}$ point with $P$ being the total number of points
while $t \in [0, T]$ refrs to the frame index, with $T$ denoting the number of frames in the video.
We refer to a point at time step $t$ as $p_t^i \in \mathbb{R}^{2 } $.
The set of all point tracks within a video is denoted as $\mathbf{P}$.

\subsection{Network Architecture}

\begin{figure}[ht]
    \centering
    \begin{subfigure}{\textwidth}
        \includegraphics[width=\hsize]{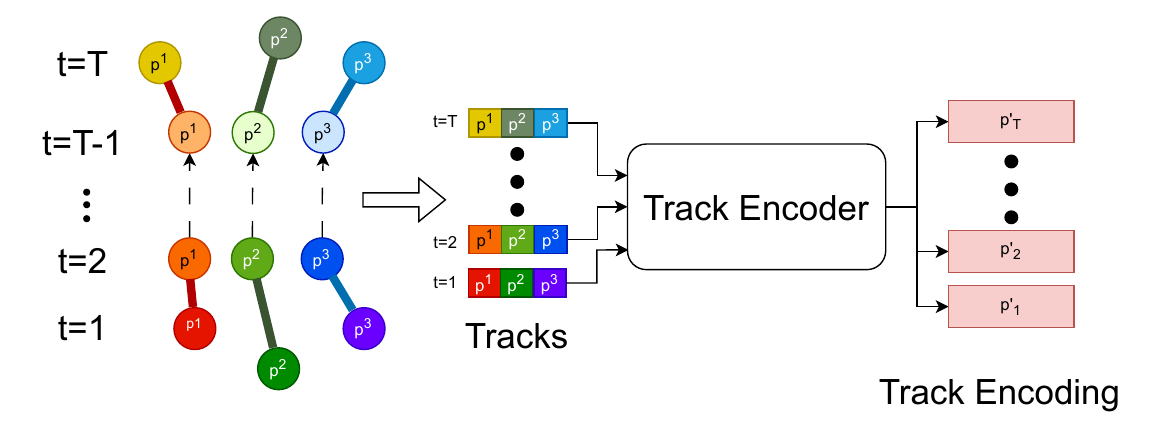}
        \caption{ }
        \label{fig:track_encoder}
    \end{subfigure}
\hfill
    \begin{subfigure}{0.325\textwidth}
    \includegraphics[width=\hsize]{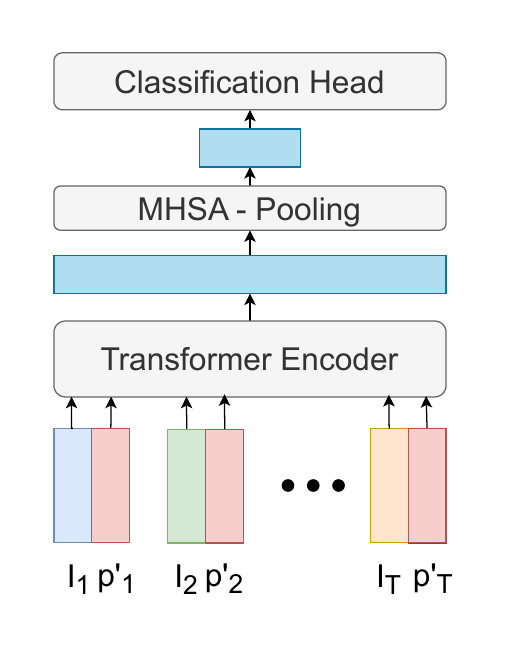}
    \caption{ }
    \label{fig:trec}
\end{subfigure}
    \caption{Network Architecture}
    \label{fig:network_arch}
\end{figure}

An overview of the proposed network architecture is shown in Figure~\ref{fig:network_arch}. 
The input to the network is a video $v \in \mathbb{R}^{T \times H \times W \times 3}$ where $H, W$ correspond to the image height and width, respectively.
Additionally, the model receives 2D point tracks $\mathbf{P}$ extracted from the video. 
The architecture consists of an image encoder $I(v_t)$ with $v_t \in \mathbb{R}^{h \times w \times 3}$ as backbone and a transformer module $G:(i, p) \rightarrow g$ which takes the image features $i$ extracted by the encoder and the point tracks $t$ as input.
The output $g$ of the transformer is pooled by an additional multi-head attention layer $M: g \rightarrow o$ which aggregates information from all tokens. 
Finally, the class token is passed through an MLP-based classification head to predict the action label.

\subsection{Action Recognition with Transformer}

The transformer model receives as input the image features extracted by the encoder together with the corresponding 2D point tracks. 
This joint representation allows the model to integrate appearance and motion information in a unified latent space. 
Each frame contributes both spatial features from the image encoder and geometric motion cues from the tracked points. 
By processing these modalities jointly, the transformer learns to associate appearance changes with underlying motion dynamics, enabling robust recognition of fine-grained actions.

Within the transformer, spatio-temporal dependencies are modeled through self-attention across both frame and point dimensions. 
This mechanism allows the network to emphasize motion patterns most relevant for action discrimination, encompassing subtle object deformations and coordinated hand movements, without requiring explicit hand or object detection.
The multi-head attention pooling layer then aggregates the sequence of token embeddings into a single representation guided by the class token. 
The resulting feature vector encodes both global context and localized motion cues relevant to the observed activity.

Finally, the pooled class token is passed through a lightweight MLP classification head to produce the final action prediction. 
During training, the model is optimized using a standard cross-entropy loss over the action classes. 
This simple yet effective design allows TRec to capture meaningful motion representations directly from tracked points and image features, achieving competitive performance while maintaining low computational complexity.

\section{Experiments}

We conduct a series of experiments to evaluate the effectiveness of the proposed model under various configurations. 
All models are implemented using a ResNet18 backbone combined with a Transformer architecture. 
The Transformer module consists of six attention layers with four heads each and an intermediate dimension of 2048. 
Before being fed into the Transformer, the 2D point tracks $p$ are normalized and reshaped to $(B, T \times 2, P)$, then passed through an MLP to project them to the Transformer's input dimensionality. Here, $B$ denotes the batch size.
The number of frames per video is set to 8 if not stated otherwise. 
The number of point tracks is randomized between 200 and 400 points for each mini-batch.

We compare two model variants: 
(1) a baseline using only RGB input, 
(2) a trackpoint model that incorporates 2D point trajectories

The baseline receives only the video frames as input.
Unless otherwise stated, training is performed with a batch size of 32, AdamW optimizer~\cite{loshchilovDecoupledWeightDecay2019b}, 
a learning rate of $1\times10^{-4}$ for the Transformer, and $1\times10^{-5}$ for the backbone network. 
We use a cosine warm-restart learning rate scheduler with 5000 warm-up iterations 
and train for 20 epochs on an NVIDIA A40 GPU. 
Input frames are resized to $256\times256$ and augmented using random cropping, 
horizontal flipping, Gaussian blur, and color jitter. The tracks are agumented according to the augmentation of their corresponding image.
The observation horizon is set to $8$ frames. 
Out of the 900 tracked points generated per video, 400 points are randomly selected and provided to the network 
to promote diversity.
The random number generators are seeded in all tests, thus in each experiment the tested model receives the same point tracks. 

\subsection{Track vs No Track}

The first experiment compares the performance of TRec with a baseline model that shares the same architecture and training parameters but only receives the RGB frames as input. 
The results, summarized in Table~\ref{tab}, demonstrate that TRec achieves substantially higher accuracy than the baseline. 
This improvement indicates that the model effectively leverages the 2D point tracks to capture motion dynamics and accurately infer the actions occurring in the video.

\begin{table}
    \caption{Performance comparison with and without point tracks}
    \label{tab}
    \centering
    \begin{tabular}{ | c | c | c | }
        \hline
        Model & Top-1 & Top-5 \\
        \hline
        TRec & $61.10\pm8.66$ & $83.95\pm6.62$ \\
        baseline & $30.27\pm 8.05$ & $53.24\pm8.75$ \\
        \hline
    \end{tabular}
\end{table}

\subsection{Number of points}

\begin{figure}[h!]
    \centering
    \includegraphics[width=\hsize]{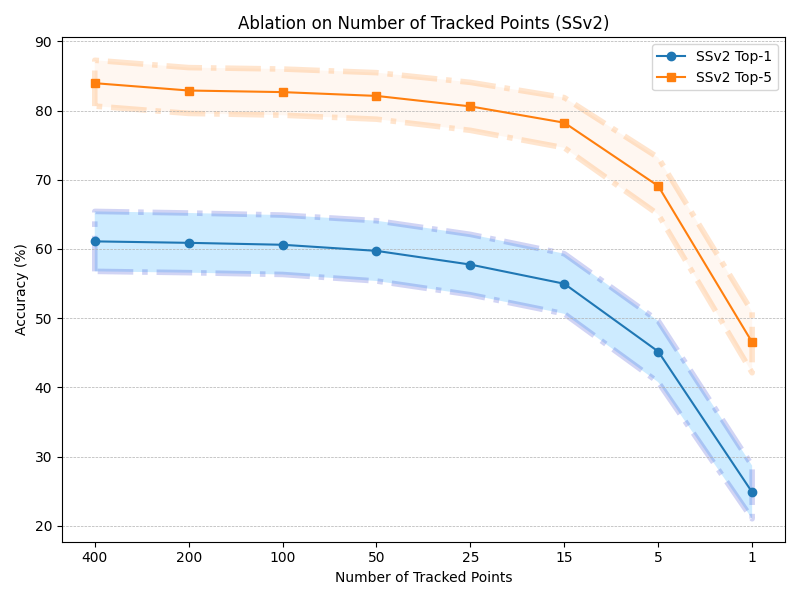}
    \caption{ Ablation on the number of point tracks used as input to the model.}
    \label{fig:ablation_num_points}
\end{figure}

To assess the influence of the number of tracked points on action recognition performance, 
we conduct an ablation study by varying the number of input points provided to the model. 
This experiment investigates how sensitive TRec is to the density of motion information encoded in the point trajectories. 
The quantitative results are summarized in Table~\ref{tab:numframes}, 
while Figure~\ref{fig:ablation_num_points} illustrates how performance changes with different numbers of points. 
The model maintains stable accuracy when using 50 or more tracked points, 
with diminishing performance gains beyond 100 points. 
However, when fewer than 25 point tracks are used, the model's accuracy drops noticeably, 
indicating that a minimal number of trajectories is required to capture sufficient motion cues 
for reliable action recognition.

\begin{table}[h!]
    \caption{Ablation on the number of tracked points}
    \label{tab:numframes}
    \centering
    \begin{tabular}{ | c | c | c |}
        \hline
        Number of Points & Top-1 & Top-5 \\
        \hline
        400 &  \textbf{61.10 $\pm$8.66 } & \textbf{83.95 $\pm$6.62 } \\
        \hline
        200 &  60.89 $\pm$8.62  & 82.89 $\pm$6.62  \\
        \hline
        100 &  $60.61 \pm8.58 $ & $82.66 \pm6.67 $ \\
        \hline
        50 &  $59.73 \pm8.71 $ & $82.11 \pm6.72 $ \\
        \hline
        25 &  $57.75 \pm8.72 $ & $80.60 \pm6.93 $ \\
        \hline
        15 &  $54.98 \pm8.68 $ & $78.24 \pm7.25 $ \\
        \hline
        5 &  $45.19 \pm8.85 $ & $69.10 \pm8.20 $ \\
        \hline
        0 &  $24.87 \pm7.68 $ & $46.53 \pm8.82 $ \\
        \hline
    \end{tabular}
\end{table}

\subsection{Background Motion as Context for Action Recognition}

To investigate the influence of background motion on action recognition, 
we analyze the effect of filtering point tracks using a Kernel Density Estimation (KDE) approach. 
Our main hypothesis is that training the model on filtered points, which primarily represent hands and manipulated objects, will lead to slightly lower performance compared to using all points, 
as background motion provides valuable contextual cues for recognizing actions.

\begin{figure}[ht]
    \centering
    \captionsetup[subfigure]{labelformat=empty}
    \begin{subfigure}{0.325\textwidth}
        \includegraphics[width=\hsize]{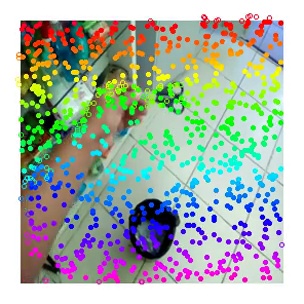}
        \caption{ }
    \end{subfigure}
\hfill
    \begin{subfigure}{0.325\textwidth}
        \includegraphics[width=\hsize]{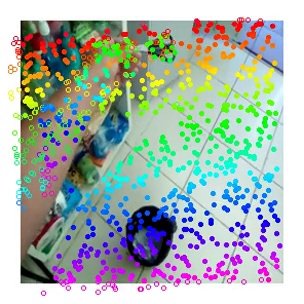}
        \caption{ }
    \end{subfigure}
\hfill
    \begin{subfigure}{0.325\textwidth}
        \includegraphics[width=\hsize]{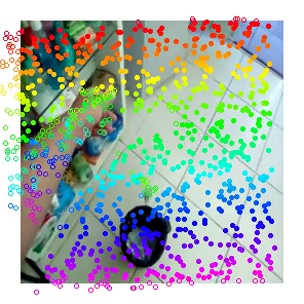}
        \caption{ }
    \end{subfigure}
    \caption{Points without filtering}
    \centering
    \captionsetup[subfigure]{labelformat=empty}
    \begin{subfigure}{0.325\textwidth}
        \includegraphics[width=\hsize]{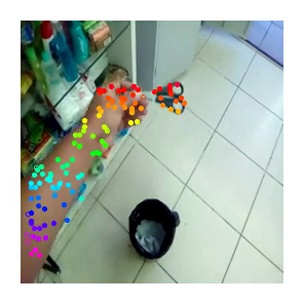}
        \caption{ }
    \end{subfigure}
    \begin{subfigure}{0.325\textwidth}
        \includegraphics[width=\hsize]{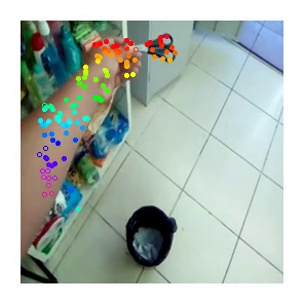}
        \caption{ }
    \end{subfigure}
    \begin{subfigure}{0.325\textwidth}
        \includegraphics[width=\hsize]{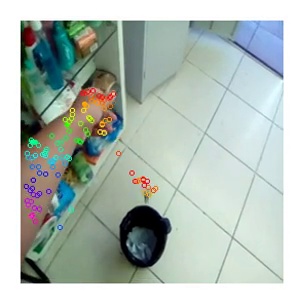}
        \caption{ }
    \end{subfigure}
    \caption{Points filtered using KDE}
    \label{fig:filtered_point_tracks}

\end{figure}

Filtering is performed by applying KDE to the motion vectors of the tracked points, 
as illustrated in Figure~\ref{fig:filtered_point_tracks}. 
Since most points in the image belong to the background and therefore share similar motion patterns, 
the KDE identifies these high-density motion clusters corresponding to background movement. 
By excluding these clusters, we retain mainly the low-density points associated with hands and active objects. 
We then train a model using only the filtered points and compare its performance to a model trained on the full set of tracks that includes both foreground and background motion. 
The model trained only on filtered points is denoted here as FilterTRec.
The quantitative results of this comparison are presented in Table~\ref{tab:filter}.

The standard TRec model achieves a Top-1 accuracy of $61.10\%$. 
However, when the input points are filtered using KDE, performance drops notably to $52.46\%$. 
This decline indicates that the model effectively utilizes background motion cues and partially relies on them for action prediction. 
To further validate this observation, we trained a variant of our model exclusively on the filtered points. 
As shown in Table~\ref{tab:filter}, the performance decreases even further (Top-1 accuracy of $36.34\%$), confirming that background motion contributes significantly to the model's recognition capability.

\begin{table}[h!]
    \caption{Model evaluation with KDE}
    \label{tab:filter}
    \centering
    \begin{tabular}{ | c | c | c |}
        \hline
        \multirow{2}{*}{Model} & \multicolumn{2}{c|}{SSv2} \\
        \cline{2-3}
        & Top-1 & Top-5 \\
        \hline
        vanilla TRec & $61.10 $ & $83.95 $ \\
        vanilla TRec + KDE & $52.46 $  & $74.04 $  \\
        \hline
        FilterTRec + KDE & $36.34 $ & $60.57 $ \\
        \hline
    \end{tabular}
\end{table}

\subsection{Single-Image Evaluation}

To further analyze the contribution of the tracked points, we train and evaluate the model using only a single image while retaining the complete set of point tracks from the corresponding video. 
This experiment isolates the influence of motion trajectories from visual appearance and emphasizes the model's ability to infer actions purely based on the temporal evolution of the tracked points. 

In this setup, only the initial image of each video is provided as input during training and testing. 
Because this frame may not contain visible hands or objects, the model must rely entirely on the tracked motion to infer hand movements and object interactions. 
Moreover, the Something-Something-v2 dataset includes several action categories that can only be recognized by observing the final frames of a video, such as \textit{pretending to do something} or \textit{revealing something}. 
As the model cannot predict these actions without the final frame, we evaluate the approach on the original dataset and on a version excluding ambiguous categories whose annotations contain \textit{pretend}, \textit{uncover}, \textit{reveal}, or \textit{fail}.

As shown in Table~\ref{tab:single_img}, the model achieves higher performance than the RGB-only baseline, demonstrating that the learned motion representations from the point tracks provide strong cues for action recognition even when only a single frame is available.
As input we only provide the initial image. That image might not contain the hands, thus the model has to extract the hand motions from the point tracks.

\begin{table}[h!]
    \centering
    \caption{Single Image Evaluation}
    \label{tab:single_img}
    \begin{tabular}{  | c | c | c |}
        \hline
        Model & Top-1 & Top-5 \\
        \hline
        TRec &  44.7  & 69.12  \\
        baseline & 28.53  & 51.15  \\
        \hline
    \end{tabular}

    \bigskip
    Performance on the Something-Something-v2 dataset using initial image as input and additionaly the point tracks for TRec.
\end{table}

\section{Discussion}

The experimental results demonstrate that incorporating 2D point tracks provides a clear advantage for egocentric action recognition compared to purely RGB-based methods. 
By explicitly encoding motion cues through tracked points, our model captures spatial-temporal dynamics that RGB alone cannot represent. 
This finding supports our initial hypothesis that motion trajectories offer a lightweight and interpretable alternative to computationally demanding approaches such as optical flow or hand-object detection.

The ablation on the number of tracked points indicates that the model is robust to variations in track density. 
Performance remains stable with as few as 50 points, and improvements beyond 100 points become marginal. 
This suggests that the model efficiently extracts the most relevant motion information without requiring dense tracking, offering a favorable trade-off between computational cost and accuracy. 
However, performance degrades substantially when fewer than 25 points are used, implying that a minimal level of motion diversity is necessary for reliable recognition.

The single-image experiment further highlights the discriminative power of the point trajectories. 
Even when only the first frame of a video is available, the model achieves higher accuracy than the RGB-only baseline, demonstrating that temporal motion encoded by the tracks carries sufficient information to infer the underlying action. 
This result also underlines the interpretability of our representation, in which the model can reason about hand and object movement even when they are not visible in the initial frame.

Finally, the kernel density filtering experiment reveals that background motion contributes significantly to action recognition in egocentric videos. 
Removing background points leads to a noticeable drop in accuracy, suggesting that the model relies on global motion patterns, encompassing camera and head movements, as contextual cues for action understanding.
This aligns with the egocentric nature of the data, where background dynamics often correlate with the actor's motion and intention.
 
Overall, these findings confirm that point-track representations provide an effective and computationally efficient way to model motion in egocentric videos. 
Future work could focus on refining the integration of point-based and semantic cues, dynamically weighting background and foreground contributions, 
and developing improved approaches for collecting and maintaining stable track points in egocentric scenarios with substantial head movement, where existing tracking methods often degrade in performance.
\bibliographystyle{splncs04}
\bibliography{bibliography.bib}
\end{document}